\documentclass[review]{elsarticle}
\usepackage{multirow}
\usepackage{array,balance}
\usepackage{graphicx}
\usepackage{amsmath}
\usepackage{amssymb}
\usepackage{color}

\usepackage{lineno,hyperref}

\journal{Neurocomputing}

\begin{document}

\begin{frontmatter}

\title{An Unsupervised Domain Adaptation Model based on Dual-module Adversarial Training}

\author[mymainaddress]{Yiju Yang}
\author[mymainaddress]{Tianxiao Zhang}
\author[mymainaddress]{Guanyu Li}
\author[mymainaddress]{Taejoon Kim}
\author[mysecondaryaddress]{Guanghui Wang\corref{mycorrespondingauthor}}
\cortext[mycorrespondingauthor]{Corresponding author}
\ead{wangcs@ryerson.ca}

\address[mymainaddress]{Department of Electrical Engineering and Computer Science, University of Kansas, Lawrence, KS, USA, 66045}
\address[mysecondaryaddress]{Department of Computer Science, Ryerson University, Toronto, ON, Canada M5B 2K3}

\begin{abstract}
In this paper, we propose a dual-module network architecture that employs a domain discriminative feature module to encourage the domain invariant feature module to learn more domain invariant features. The proposed architecture can be applied to any model that utilizes domain invariant features for unsupervised domain adaptation to improve its ability to extract domain invariant features. We conduct experiments with the Domain-Adversarial Training of Neural Networks (DANN) model as a representative algorithm. In the training process, we supply the same input to the two modules and then extract their feature distribution and prediction results respectively. We propose a discrepancy loss to find the discrepancy of the prediction results and the feature distribution between the two modules. Through the adversarial training by maximizing the loss of their feature distribution and minimizing the discrepancy of their prediction results, the two modules are encouraged to learn more domain discriminative and domain invariant features respectively. Extensive comparative evaluations are conducted and the proposed approach outperforms the state-of-the-art in most unsupervised domain adaptation tasks.
\end{abstract}


\end{frontmatter}


\section{Introduction}
In recent years, deep convolutional neural networks (CNNs) have achieved huge success in many computer vision applications, such as image classification \cite{ cen2021deep, li2021sgnet}, object detection \cite{ li2021colonoscopy, ma2018mdcn}, tracking \cite{zhang2020real}, segmentation \cite{ he2021sosd, patel2021enhanced}, image generation \cite{xu2021drb}, and crowd counting \cite{ sajid2021multi}. A lot of powerful network architectures have been proposed for efficient feature extraction, representation, and optimization \cite{he2016deep, xu2020towards, zhang2018bpgrad}. However, most of these models need to be trained via supervised learning from a large collection of reliable training data.

In many practical application scenarios, we may not have or it is too expensive to obtain a lot of reliable labeled data, and labeling data correctly and accurately is usually more difficult than collecting data. For example, in medical image processing, labeling data correctly requires people to have sufficient professional training and domain knowledge. Therefore, how to make full use of unlabeled data becomes a more and more attractive topic in the computer vision field. An effective method to solve this kind of problem is to use unsupervised domain adaptation (UDA).
  
In unsupervised domain adaptation, we assume that there are two data sets. One is an unlabeled data set from the target task, called the target domain. The other data set is a labeled data set from the source task, called the source domain. There are some similarities between the two data sets, however, there are still some differences between them. For example, cartoon characters and real human beings have some commonalities, but we can also clearly distinguish the differences between them. When we train a model with the source domain data set, and then apply the trained model to the source task and the target task respectively, the performance for the target domain is usually significantly lower than that for the source domain. This is due to the domain shift between the source domain and the target domain. The extent of these reductions depends on the domain shift. If the domain shift is larger, the performance loss for the target task becomes greater. To solve this problem, we need to know how to align the source domain with the target domain and reduce the shift between the two domains.
  
With the advent of gradient reversal layers (GRL) from Domain-Adversarial Training of Neural Networks (DANN) \cite{ganin2015unsupervised}, more and more people realize that adversarial training has a significant effect on aligning the feature of the source and target domains. DANN \cite{ganin2015unsupervised} extracts the global invariant features by training a domain discriminator to fool the feature extractor.  The domain discriminator is a component composed of several fully connected layers. Its function is to distinguish the input data from the source domain or the target domain. If the discriminator could not recognize the extracted feature map, it means that the extracted feature map comes from the common space of these two domains. 
The global invariant features are from this space, and DANN 
  utilizes a classifier for adversarial training to ensure the effectiveness of the features learned by the network. In addition to the adversarial training method based on GRL, there are many other adversarial training methods based on generative adversarial nets (GANs) \cite{NIPS2014_5423, liu2016coupled, disentangled-representation-learning-gan-for-pose-invariant-face-recognition, Hu_2018_CVPR, gui2020review}. Most of them have one thing in common, i.e., they adapt to the target domain by learning global invariant features.
    \begin{figure}[tp!]
    \begin{center}
        \includegraphics[scale=0.7]{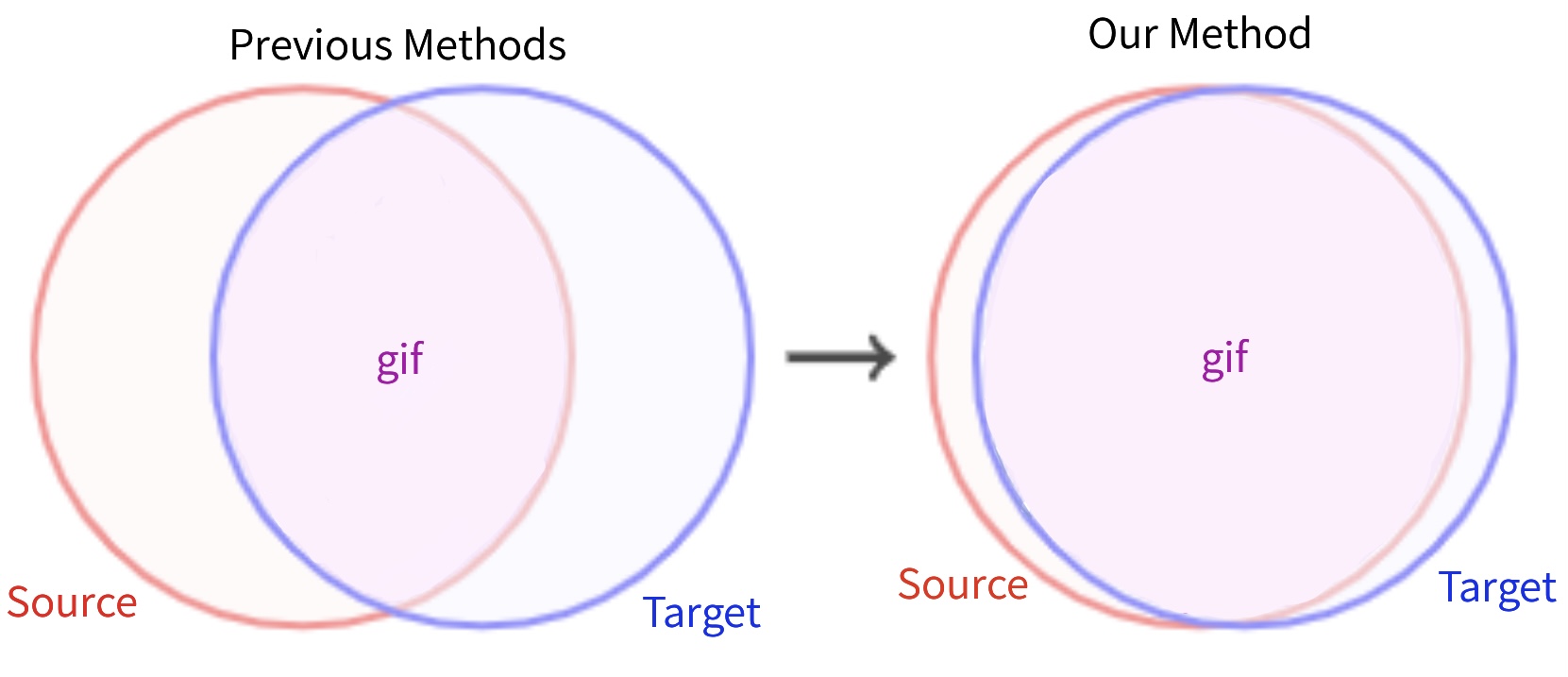}
    \end{center}
    \caption{The red circle represents the features extracted from the source domain, and the blue circle denotes the features extracted from the target domain.
    The intersection area of the two circles is the gif (global invariant feature).
    Our method can further extract global invariant features on the basis of previous methods, reducing the distance between the source domain and the target domain.}
    \label{fig:pic1}
    \end{figure}
    
  \textbf{Motivation.} Taking DANN \cite{ganin2015unsupervised} as an example, although it has the advantage to use a discriminator to encourage the model to learn invariant features, there is also a bottleneck. When we only rely on the discriminator to control the extraction ability of the domain invariant features, the conditions to extract domain invariant features will become very limited. When the extracted domain invariant features are strong enough to fool the discriminator, it will become difficult for the model to further improve its ability to extract the domain invariant features.
  In addition, when we only use the source domain labeled data to adjust the classifier, the extracted features will be more biased towards the source domain, which further limits its performance on the target task. This challenge also exists in other methods that employ the domain discriminator to extract the domain invariant features.
  
  \textbf{Contributions.}  In order to further improve the extraction ability of domain invariant features, we propose a dual-module architecture to solve this challenge. As shown in Figure \ref{fig:pic1},
  The proposed network is composed of a discriminative feature learning module and a domain invariant feature learning module, and the domain discriminative feature module is employed to encourage a domain invariant feature module to learn more domain invariant features.
  We take DANN \cite{ganin2015unsupervised} as an example to implement our model. We also employ the Maximum Classifier Discrepancy (MCD) \cite{saito2018maximum} to learn the domain discriminative features from the target domain so that the classifier is not only affected by the domain discriminative features of the source domain. 
  
  In summary, the main contributions of this paper include: 
\begin{enumerate}
\setlength{\itemsep}{-2pt}
  \item We propose a novel dual-module network architecture to promote the learning of domain invariant features;
  \item We propose a new adversarial loss to maximize the discrepancy of the feature distributions while minimizing the discrepancy of the prediction results;
  \item We propose to maximize the classifier discrepancy to ameliorate the training imbalance issue of previous unsupervised domain adaptation methods.
\end{enumerate} 
  
We evaluated our proposed method on four popular benchmark datasets and compared its performance with previous work. The proposed structure outperforms the state-of-the-art by a large margin on unsupervised domain adaptation of digit classification and object recognition. The source code of the proposed model can be accessed at the author's website\footnote{https://github.com/rucv/DMDA}.

\section{Related Work}

 Domain adaptation could be taken as a subset of transfer learning \cite{pan2009survey}, which is a commonly used technique in many computer vision tasks to improve the generalization ability of a model trained on a single domain. In this section, we describe some existing domain adaptation methods.

 {\bf Learn domain invariant features.}
 Recently, \cite{chen2020simple} explored what enables the contrastive prediction tasks to learn useful representations.
 \cite{Carlucci_2019_CVPR} learns the semantic labels in a supervised fashion, and broadens its understanding of the data by learning from self-supervised signals how to solve a jigsaw puzzle on the same images. DICD \cite{li2018domain} proposes to learn domain invariant features and class discriminative features simultaneously by a low-dimension latent feature space so that the samples with the same category are close to each other while the samples with different categories are away from each other.

 {\bf Distance-based methods.}
 Aligning the distribution between the source domain and the target domain is a very common method in solving unsupervised domain adaptation problems. Maximum Mean discrepancy (MMD) \cite{gretton2012kernel,long2017deep,tzeng2014deep,long2015learning} is a method of measuring the difference of two distributions. 
 DAN \cite{long2015learning} explored the multi-core version of MMD to define the distance between two distributions. 
 JAN  \cite{long2017deep} learned a transfer network by aligning the joint distributions of multiple domain-specific layers across the domains based on a joint maximum mean discrepancy (JMMD) criterion.
  \cite{long2016unsupervised} enabled the classifier adaptation by plugging several layers into the deep network to explicitly learn the residual function with reference to the target classifier.
 CMD \cite{zellinger2017central} is a metric on the set of probability distributions on a compact interval.
 
 To solve the problem of unbalanced datasets, Deep Asymmetric Transfer Network (DATN)  \cite{wang2018deep} proposed to learn a transfer function from the target domain to the source domain and meanwhile adapting the source domain classifier with more discriminative power to the target domain.
 DeepCORAL \cite{sun2016deep} builds a specific deep neural network by aligning the distribution of second-order statistics to limit the invariant domain of the top layer. 
  \cite{chen2020homm} proposed a Higher-order Moment Matching (HoMM) method to minimize the domain discrepancy.

 {\bf Adversarial methods.}
 Adversarial training is another very effective method to transfer domain information. 
 Inspired by the work of gradient reversal layer  \cite{ganin2015unsupervised}, a group of domain adaptation methods has been proposed based on adversarial learning. 
 RevGrad  \cite{ganin2015unsupervised} proposed to learn the global invariant feature by using a discriminator that is used to reduce the discriminative features in the domain.
 Deep Reconstruction-Classification Networks (DRCN)  \cite{ghifary2016deep} jointly learned a shared encoding representation for supervised classification of the labeled source data, and unsupervised reconstruction of the unlabeled target data.
  \cite{bousmalis2016domain} extracted image representations that are partitioned into two subspaces. Adversarial Discriminative Domain Adaptation (ADDA)  \cite{tzeng2017adversarial} trained two feature extractors for the source and target domains respectively, to generate embeddings to fool the discriminator.
  
 Maximum Classifier Discrepancy (MCD) \cite{saito2018maximum,yang2021multiple} proposed to explore task-specific decision boundaries. 
 CyCADA \cite{hoffman2018cycada} introduced a cycle-consistency loss to match the pixel-level distribution. 
 SimeNet  \cite{pinheiro2018unsupervised} solved this problem by learning the domain invariant features and the categorical prototype representations.
 CAN  \cite{kang2019contrastive} optimized the network by considering the discrepancy of the intra-class domain and the inter-class domain.
 Graph Convolutional Adversarial Network (GCAN)  \cite{ma2019gcan} realized the unsupervised domain adaptation by jointly modeling data structure, domain label, and class label in a unified deep model.
  \cite{gong2019dlow} proposed a domain flow generation (DLOW) model to bridge two different domains by generating a continuous sequence of intermediate domains flowing from one domain to the other.
  \cite{cai2019learning} employed a variational auto-encoder to reconstruct the semantic latent variables and domain latent variables behind the data.
 Drop to Adapt (DTA)  \cite{lee2019drop} leveraged adversarial dropout to learn strongly discriminative features by enforcing the cluster assumption.
 Instead of representing the classifier as a weight vector,  \cite{lu2020stochastic} modeled it as a Gaussian distribution with its variance representing the inter-classifier discrepancy.

 

DMRL \cite{wu2020dual} utilizes dual mixup regularized learning to improve the class-level prediction in the target domain and boost domain invariant feature extracting. DMRL \cite{wu2020dual} focuses on data augmentation by generating extra mixed samples to improve the performance of adversarial domain adaptation. Dual modules have demonstrated great effectiveness with adversarial learning in unsupervised domain adaptation. DADA \cite{du2020dual} is the first model which introduces a dual mechanism with adversarial learning in domain adaptation. It utilizes two discriminators whose outputs include classification predictions from both source and target domains so that domain-level and class-level adaptation could be considered at the same time in each discriminator. Although there are two discriminators in the model, they share the feature extractor. In the proposed model, the two discriminators have separate feature extractors that are focused on domain invariant features and domain discriminative features, respectively. Different from DMRL and DADA, we designed a new network structure for adversarial domain adaptation, and the proposed model executes the adversarial learning in the M1 module and maximizes the discrepancy of feature distributions between the two modules.

\section{Proposed Method}
 In this paper, we consider the unsupervised domain adaptation problem.
 Suppose we have a source domain $D_s = {\{(X_s, Y_s)\}} = {\{(x_s^i, y_s^i)\}}_{i=1}^{n_s}$ with $n_s$ labeled samples and a target domain $D_t = {\{(X_t)\}} = {\{(x_t^i)\}}_{i=1}^{n_t}$ with $n_t$ unlabeled samples.
The two domains share the same label space $Y = \{1,2,3,...,K\}$, where $K$ is the number of categories. We assume that the source sample $x_s$ belongs to the source distribution $P_s$, and the target sample $x_t$ belongs to the target distribution $P_t$, where $P_s \neq P_t$. 
Our goal is to train a classifier $f_\theta (x)$ that can minimize the target risk $\epsilon _t = E_{x\in D_t}[f_\theta (x) \neq y_t]$, where $f_\theta (x)$ represents the output of the deep neural network, and $\theta$ represents the model parameters to be learned.

In the following subsections, we will present the main idea of the proposed strategy for unsupervised domain adaptation, the network architecture, and the training steps of the network.
 
 
\subsection{Main Idea}
\label{ss:1}
In the method that employs the domain discriminator to help the feature extractor learn domain invariant features, the space represented by the domain invariant features is a fusion of the source domain and the target domain.
We believe that the ideal state of this type of method is to first focus on learning the domain invariant features, and then learn some domain discriminative features from the target domain until the process reaches its upper limit to further improve the performance.
When using the domain discriminator as one of the conditions of adversarial training, a big issue is that, with the increase of the model's ability to learn domain invariant features, the ability of the discriminator to give correct judgment to the data will be greatly reduced. When it decays beyond a certain threshold, the model will not be able to further improve the learning ability of domain invariant features. Thus, the domain discriminator has become a bottleneck of learning domain invariant features, restricting the model to reach its upper limit of learning domain invariant features.
 
\begin{figure*}[tp]
\begin{center}
  \includegraphics[width=12cm, height=8cm]{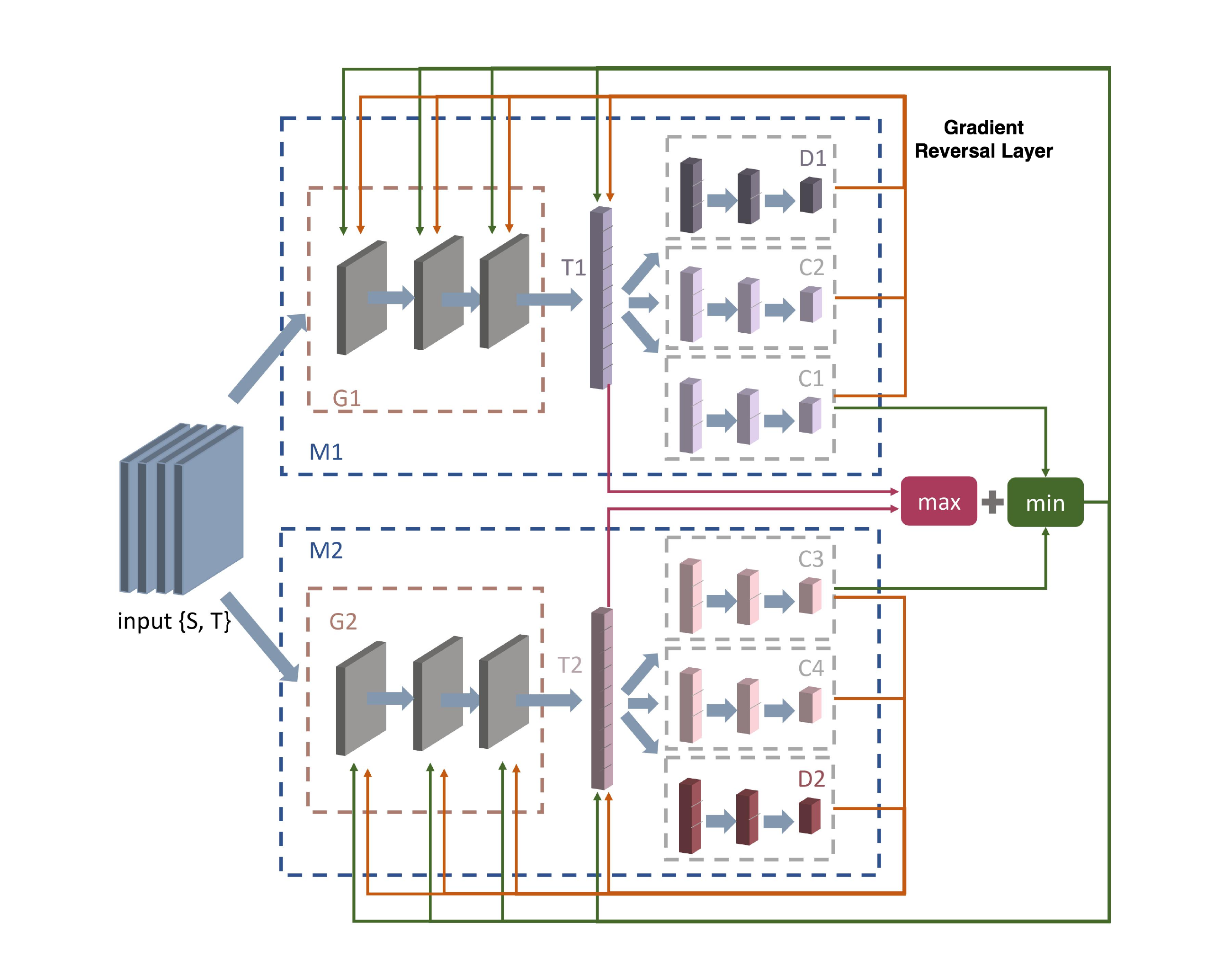}
\end{center}
  \caption{The proposed network architecture has two modules. M1 learns the domain invariant features, and M2 learns the domain discriminative features. G1 is the feature extractor of M1, and G2 is the feature extractor of M2. T1 and T2 are learner transformation layers. D1 and D2 are domain discriminators. C1, C2, C3, and C4 are classifiers. Orange lines denote the backward process of training steps 1 and 2, and Green lines denote the backward process of training step 3. The gradient reversal layer in M1 helps the feature extractor to learn domain invariant features, while M2 does not have the gradient reversal layer so that the feature extractor can learn the domain discriminative features.}
\label{fig:long}
\end{figure*}

 To address 
 this challenge, we propose a dual-module structure to further improve the domain invariant feature extraction capabilities of the original algorithm.
 As shown in Figure \ref{fig:long}, one module is designed to learn the domain-invariant features in the data, and the other module is used to learn the domain discriminative features in the data. Then we encourage the two modules to their extremity by constructing a new adversarial loss function. 
 As a result, one module learns more domain-invariant features, and the other module learns more domain discriminative features. We implement the proposed strategy on top of the most representative algorithm DANN \cite{ganin2015unsupervised} that employs a domain discriminator to learn domain invariant features. In the adversarial training of DANN, only the source data is used to adjust the classifier. Therefore, we adopt Maximum Classifier Distance (MCD)  \cite{saito2018maximum} to learn the domain discriminative features from the target domain after the model reaches its upper limit to further improve its performance.
 
 \textbf{Dual-module architecture.}
 In the dual-module architecture, we need to build two modules with completely different functions. As shown in Figure \ref{fig:long}, We call the module that can learn the domain invariant features as M1, and the module that can learn domain discriminative features as M2. Different from \cite{ding2018semi} which employs coupled neural networks for the source domain and target domain respectively, both modules in our model receive the data from the source domain and the target domain since each module has a discriminator.
 For the module M1, we can use any algorithm that uses a discriminator to learn domain invariant features. In our experiment, we use DANN \cite{ganin2015unsupervised} as the module M1, since it has a clear and concise structure. For the module M2, the easiest way is to employ the source-only method. As the number of training increases, the ability of the feature extractor to extract domain discriminative features will be greatly improved.
 In other words, among the features extracted by M2, the domain discriminative features will have a larger ratio than the domain invariant features.
 
 In our experiments, our M2 also follows the same DANN structure. 
 The only difference is that we remove the gradient reversal layer for adversarial training, so the discriminator will make the learned features more discriminative.
 As a result, when we input the same data into the two modules, the feature distribution from M1 will be based on domain invariant features, and the feature distribution from M2 will be based on domain discriminative features.
 We designed an adversarial loss for training this dual-module architecture.
 In the adversarial loss, we expand the discrepancy in feature distribution between the two modules. 
 At the same time, we narrow the discrepancy in the prediction between the two modules.
 We will provide more details about this in Section \ref{ss:3}.

 \textbf{Target domain discriminative feature.}
 Maximum Classifier Discrepancy (MCD)  \cite{saito2018maximum} is to make the model learn the decision boundary of a specific task by maximizing the prediction difference of the two classifiers. 
 Since the two classifiers are iterated under the same conditions, their prediction results are very similar. 
 However, for some data that is easy to be confused, the results given by the two classifiers may be different.
 Therefore, if the information captured by the feature extractor is more discriminative, the features that determine the decision boundary will also be larger.
 We follow this idea and use MCD to optimize our classifier, since the classifier of DANN is only adjusted by the source domain.
 We add a classifier to each of the two modules, which means that there are two classifiers with the same structure in each module.
 We assign these two classifiers to random parameters and train them together. 
 We make our module learn the domain discriminative features from the target domain by learning the decision boundary. 
 This can make up for the lack of the domain discriminative features only from the source domain.


\subsection{Network Structure}
\label{ss:2}
 In this section, we will elaborate on the proposed network structure in detail. 
 As shown in Figure \ref{fig:long}, our network employs a dual-module structure. 
 M1 is the domain invariant feature module, and M2 is the domain discriminative feature module.
 M1 is composed of a feature extractor G1, a discriminator D1, two classifiers C1 and C2, and a Linear transformation layer T1.
 M2 consists of a feature extractor G2, a discriminator D2, two classifiers C3 and C4, and a Linear transformation layer T2.
 Since the loss of our dual-module architecture needs to calculate the discrepancy of the two feature distributions, we need an independent linear transformation layer to convert the feature map into a feature distribution.
 For each module, We embed a linear transformation layer after the feature extractor. For the linear transformation layer, we set its input and output sizes as that of the feature extractor, so as to ensure that there is no information loss in this process.
 
 In addition, the domain discriminators in these two modules have completely different functions. The discriminator D1 plays the same role as the discriminator in DANN to fuse the two domains. However, the discriminator D2 is employed to separate the two domains. Therefore, M1 learns domain invariant features, while M2 learns the domain discriminative features. Please note that the sub-components used in our two modules have exactly the same structure. 
 For example, G1 and G2 have the same structure, D1 and D2 are the same, and C1, C2, C3, and C4 are the same.
 
\subsection{Discrepancy Loss}
 In this study, we follow the discrepancy loss in  \cite{saito2018maximum}. 
 We use the absolute value of the difference between the probability outputs of the two classifiers as the discrepancy loss: 
 \begin{equation}dis(p^1,p^2) = \frac{1}{K} \sum^K_{k=1} |p^1_k - p^2_k| \end{equation}
 where $p^1$ and $p^2$ are the probability outputs of the two classifiers respectively, which are the prediction scores for all the categories, and $K$ is the number of categories, and $p^1_k$ and $p^2_k$ are the specific values of their $k_{th}$ category.
 
\subsection{Training Steps}
\label{ss:3}
 
 {\bf Step 1:}
 Our model learns the decision boundary by using MCD \cite{saito2018maximum}. 
 Learning the decision boundary is essential to learning the discriminative feature. 
 The main reason we put this as the first step is to avoid conflict between the learning process of the decision boundary and the domain invariant feature learning. 
 In order to bring our model closer to the ideal state of domain fusion, our subsequent steps can effectively reduce the redundant domain discriminative features obtained from the decision boundary, thereby reducing the negative impact of the conflict.
 Following the setting of \cite{saito2018maximum}, we fix the number of iterations to 4 for learning the boundary in target domain 
 in all our experiments.
 Since the function of the linear transformation layer T is to convert the feature map into a feature distribution, our linear transformation layer only updates the parameters when the feature extractor updates the parameters.

 {\bf Step 2:}
 In this step, we continue to train the two modules separately. 
 Taking the M1 module as an example, we use the adversarial loss of DANN for training. 
 In other words, we train the model according to the training method from the original algorithm. After this step, the two modules begin to have some differences. 
 This step is necessary for the proposed dual-module structure. For the algorithms that do not use MCD for pre-processing, this will be the first step in the entire training process.

 For M1, we conduct adversarial training through gradient reversal layer ($grl$) to learn the domain invariant features. 
 \begin{equation}
 L_C =  L_{C1}(f_{\theta}(X_s),Y_s) + L_{C2}(f_{\theta}(X_s),Y_s)
 \end{equation}
 \begin{equation}
 L_{M1} = L_C + grl(L_{D1}(f_{\theta}(X_s))) + grl(L_{D1}(f_{\theta}(X_t)))  
 \end{equation}
 where $L_{M1}$ is the total loss of the whole M1 module, $L_C$ is the total loss of two classifiers, $L_{C1}$, $L_{C2}$, and $L_{D1}$ are the cross-entropy loss for the classifiers and discriminator. 
 
M2 does not have the gradient reversal layer, so the discriminator will prompt the feature extractor to learn the domain discriminative features. 
 \begin{equation}
 L_C =  L_{C3}(f_{\theta}(X_s),Y_s) + L_{C4}(f_{\theta}(X_s),Y_s)
 \end{equation}
 \begin{equation}
 L_{M2} = L_C + L_{D2}(f_{\theta}(X_s)) + L_{D2}(f_{\theta}(X_t))    
 \end{equation}
  where $L_{M2}$ is the total loss of the whole M2 module, $L_C$ is the total loss of the two classifiers, $L_{C3}$, $L_{C4}$, and $L_{D2}$ are the cross-entropy loss for the classifiers and discriminator. 
 
 {\bf Step 3:}
In this step, we conduct an adversarial loss function $L$ for our two modules. The two modules have separate parameters and they do not share parameters with each other. Thus, we intend to maximize the discrepancy of the features extracted by the feature extractors while minimizing the difference of the predicted classification results so that the feature extractors could concentrate on invariant features and discriminative features respectively, while the predicted results would not have much difference.
We input the same set of data into the two modules and extract the output from the transformation layer T and the classifier C.
 We use the linear transformation layer to convert the feature maps generated by the feature extractors into feature distributions, and calculate the discrepancy between the two modules, as illustrated in Figure \ref{fig:long}.  We use a gradient reversal layer ($grl$) to maximize the discrepancy of the feature distributions of the two modules.
 
Although there are two classifiers in each module, we only utilize C1 (from M1) and C3 (from M2) to calculate the discrepancy between the predictions of the two modules. At this time, the discrepancy of the predicted results between the two modules is minimized so that, although the feature extractors in the two modules focus on invariant features and discriminative features respectively, the predicted results would not have too much discrepancy.
 
 Our adversarial loss function is to play a Min-Max game with these two discrepancies.
 
 \begin{equation}
 grl(dis(t)) = \max ( dis(t_1^s,t_2^s)+dis(t_1^t,t_2^t))
 \end{equation}
 \begin{equation}
 dis(c) = \min (  dis(c_1^s,c_3^s)+dis(c_1^t,c_3^t))
 \end{equation}
 \begin{equation}
 L = grl(dis(t)) + dis(c)
 \end{equation}
  where $L$ is the total loss, $dis()$ is the discrepancy loss.
  $t_1^s$ means the output is from T1 and the input is from the source domain, and $t_2^s$ means the output from T2 and input from the source domain.
  $t_1^t$ means the output from T1 and input from the target domain, and $t_2^t$ means the output from T2 and input from the target domain.
  $c_1^s$ means the probability output from C1 and takes input from the source domain, and $c_1^t$ means the probability output from C1 and takes input from the target domain.
  $c_3^s$ means the probability output from C3 and takes input from the source domain, and $c_3^t$ means the probability output from C3 and takes input from the target domain.


\section{Experiments}
 We conducted extensive experiments to evaluate the proposed architecture and the effect of different components in the architecture. We conduct our experiments on three digits datasets, two traffic sign datasets, and one object classification dataset.
 In the following, \textbf{OURS} mentioned in the results refers to the case where only the dual-module structure is used.
 \textbf{MCD+DANN} refers to the case where MCD is directly employed to solve the imbalance problem in DANN.
 \textbf{OURS+1M} refers to the case where MCD is only used by the M1 module.
 \textbf{OURS+2M} refers to the case where MCD is used by both M1 and M2 modules.
We also compare the performance with DANN and MCD as the baselines.

\subsection{Experiments on Digits and Traffic Signs Datasets} \label{sec:exp:1}

In this section, we evaluate our model using the following five datasets: MNIST \cite{lecun-mnisthandwrittendigit-2010}, Street View House Numbers (SVHN) \cite{netzer2011reading}, USPS  \cite{291440}, Synthetic Traffic Signs (SYN SIGNS) \cite{978-3-319-02895-8_52}, and the German Traffic Signs Recognition Benchmark (GTSRB) \cite{Stallkamp-IJCNN-2011}.

\textbf{MNIST:} The dataset contains images of digits 0 to 9 in different styles. It is composed of $60,000$ training and $10,000$ testing images. 

\textbf{USPS:} This is also a digit dataset with $7,291$ training and $2,007$ testing images. 

\textbf{SVHN:} Another digit dataset with $73,257$ training, $26,032$ testing, and $53,1131$ extra training images.

\textbf{SYN SIGNS:} This is a synthetic traffic sign dataset, which contains $100,000$ labeled images, and 43 classes.

\textbf{GTSRB:} A dataset for German traffic signs recognition benchmark. The training set contains $39,209$ labeled images and the test set contains $12,630$ images. It also contains 43 classes.


We evaluate the unsupervised domain adaptation model on the following four transfer scenarios:
\begin{itemize}
\item SVHN $\xrightarrow{}$ MNIST 
\item USPS $\xrightarrow{}$ MNIST 
\item MNIST $\xrightarrow{}$USPS 
\item SYNSIG $\xrightarrow{}$ GTSRB
\end{itemize}

 \begin{table}
    \begin{center}
    \begin{tabular}{ | c |c | c | c | c | } 
    \hline
      & \textbf{\footnotesize SVHN} & \textbf{\footnotesize MNIST} & \textbf{\footnotesize USPS} & \textbf{\footnotesize SYNSIG} \\ 
      \textbf{\footnotesize Method} & {\footnotesize to} & {\footnotesize to} & {\footnotesize to} & {\footnotesize to}\\
      & \textbf{\footnotesize MNIST} & \textbf{\footnotesize USPS} & \textbf{\footnotesize MNIST} & \textbf{\footnotesize GTSRB} \\
    \hline
        {\footnotesize Source only} & {\footnotesize 67.1} &{\footnotesize 79.4} &{\footnotesize 63.4} &{\footnotesize 85.1}\\
        \hline
        {\footnotesize DANN} \cite{ganin2015unsupervised}&{\footnotesize 71.1}&{\footnotesize 85.1} &{\footnotesize 73.0$\,\pm\,$0.2} &{\footnotesize 88.7}\\ 
        {\footnotesize ADDA} \cite{tzeng2017adversarial}&{\footnotesize 76.0$\,\pm\,$1.8}& - & {\footnotesize 90.1$\,\pm\,$0.8} & -\\
        {\footnotesize CoGAN \cite{liu2016coupled}}& - & - &{\footnotesize 89.1$\,\pm\,$0.8}& -\\
        {\footnotesize PixelDA} \cite{bousmalis2017unsupervised} & - &{\footnotesize 95.9}& - & -\\
        {\footnotesize ASSC} \cite{haeusser2017associative}&{\footnotesize 95.7$\,\pm\,$1.5}& - & - &{\footnotesize 82.8$\,\pm\,$1.3}\\
        {\footnotesize UNIT} \cite{liu2017unsupervised}&{\footnotesize 90.5} &{\footnotesize 96.0}&{\footnotesize 93.6}& -\\
        {\scriptsize CyCADA \cite{hoffman2018cycada}}&{\footnotesize 90.4$\,\pm\,$0.4}&{\footnotesize 95.6$\,\pm\,$0.2}&{\footnotesize 96.5$\,\pm\,$0.1}& -\\
        {\footnotesize GTA} \cite{sankaranarayanan2018generate} &{\footnotesize 92.4$\,\pm\,$0.9}&{\footnotesize 95.3$\,\pm\,$0.7}&{\footnotesize 90.8$\,\pm\,$1.3}& -\\
        {\scriptsize DeepJDOT \cite{bhushan2018deepjdot}} &{\footnotesize 96.7}&{\footnotesize 95.7}&{\footnotesize 96.4}& -\\
        {\footnotesize SimNet \cite{pinheiro2018unsupervised}}& - &{\footnotesize 96.4}&{\footnotesize 95.6}& -\\
        {\footnotesize GICT} \cite{qin2019generatively}&{\footnotesize 98.7} &{\footnotesize 96.2}&{\footnotesize 96.6}& -\\
        {\footnotesize STAR} \cite{lu2020stochastic}&{\footnotesize 98.8$\,\pm\,$0.05} &{\footnotesize 97.8$\,\pm\,$0.1}&{\footnotesize 97.7$\,\pm\,$0.05}&{\footnotesize 95.8$\,\pm\,$0.2}\\ 
        \hline
        {\footnotesize MCD} \cite{saito2018maximum}&{\footnotesize 96.2$\,\pm\,$0.4} &{\footnotesize 96.5$\,\pm\,$0.3} &{\footnotesize 94.1$\,\pm\,$0.3} &{\footnotesize 94.4$\,\pm\,$0.3}\\
        {\footnotesize MCD+DANN} &{\footnotesize 91.4$\,\pm\,$0.2} &{\footnotesize 97.3$\,\pm\,$0.3} & {\footnotesize 96.8$\,\pm\,$0.1} &{\footnotesize 90.7$\,\pm\,$0.2}\\
        \textbf{\small ours} & {\footnotesize 98.9$\,\pm\,$0.1}&{\footnotesize 95.1$\,\pm\,$0.4}&{\footnotesize 96.1$\,\pm\,$0.2} &{\footnotesize 91.1$\,\pm\,$0.2}\\
        \textbf{\small ours+1M} &{\footnotesize 98.3$\,\pm\,$0.1} & {\footnotesize 97.1$\,\pm\,$0.2}  & {\footnotesize 97.0$\,\pm\,$0.1} & {\footnotesize 90.8$\,\pm\,$0.2}\\
        \textbf{\small ours+2M} &\textbf{\footnotesize 99.3$\,\pm\,$0.1} & \textbf{\footnotesize 98.0$\,\pm\,$0.4} & \textbf{\footnotesize 97.7$\,\pm\,$0.1} & \textbf{\footnotesize 97.0$\,\pm\,$0.2}\\
    \hline
    \end{tabular}
    \end{center}
    \caption{The performance on digit classification and sign classification. We report the mean and the standard deviation of the accuracy obtained over 5 trials.}
    \label{table:1}
\end{table}

During the experiments, we employ the CNN architecture and the input size used in  \cite{saito2018maximum}.
We used mini-batch stochastic gradient descent (SGD) to optimize our model and set the learning rate at 0.002 in all experiments. 
We follow DANN \cite{ganin2015unsupervised} and employ the SGD training schedule for the part of learning domain invariant feature: the learning rate adjusted by $\eta_p = \frac{\eta_0}{(1 + \alpha p)^ \beta}$, where $p$ denotes the process of training iterations that is normalized in [0, 1], and we set $\eta_0$ = 0.002, $\alpha$ = 10, and $\beta$ = 0.75; the hyper-parameter $\lambda$ is initialized at 0 and is gradually increased to 1 by $\lambda_p = \frac{2}{1+exp(- \gamma p)} - 1$, where we set $\gamma$ = 10. 
For the maximum classifier discrepancy, we set the hyper-parameter $k$ = 4 (the number of iterations for learning the boundary in the target domain)
in all experiments. 
We set the batch size to 128 in all experiments. 
We follow the protocol of unsupervised domain adaptation and do not use validation samples to tune the hyper-parameters. 

We present the digit classification and sign classification performance in Table \ref{table:1}.
From the table, it is clear that the proposed method outperforms previous models in all settings, where \textbf{OURS+2M} is the top-performing variant.
In order to explore the direct impact of MCD on DANN, we combined them and compared the experimental results with the state-of-the-art.
We can find that the performance of \textbf{MCD+DANN} is lower than MCD in both SVHN$\xrightarrow{}$MNIST and SYNSIG$\xrightarrow{}$GTSRB tasks. 
The result demonstrates that when MCD acts directly on DANN, it sometimes may cause conflict with DANN, while the structure of \textbf{OURS+2M} can effectively resolve this conflict.
Compared with MCD (baseline), we obtain an improvement of 3.1\% in SVHN$\xrightarrow{}$MNIST, 1.5\% in MNIST$\xrightarrow{}$USPS, 3.6\% in USPS$\xrightarrow{}$MNIST, and 2.6\% in SYNSIG$\xrightarrow{}$GTSRB. In addition, our model outperforms the state-of-the-art methods on all tasks.

\subsection{Experiments on VisDA Classification Dataset}
 We further evaluate our model on the large VisDA-2017 dataset \cite{visda2017}. 
 The VisDA-2017 image classification is a 12-class domain adaptation dataset used to evaluate the adaptation from synthetic-object to real-object images. 
 The source domain consists of 152,397 synthetic images, where 3D CAD models are rendered from various conditions. 
 The target domain consists of 55,388 real images taken from the MS-COCO dataset  \cite{lin2014microsoft}.
 
 In this experiment, we employ Resnet-18  \cite{he2016deep} as our feature extractor, and the parameters are adopted from the ImageNet pre-trained model. The pre-trained model of our Resnet-18 comes from Pytorch  \cite{paszke2017automatic} and all experimental implementations are based on Pytorch.
 
 The input images are of the size $224 \times 224$. First, we resize the input image to 256, and then crop the image to $224 \times 224$ in the center. When we train the model using only the source domain, we just modify the output size of the original last fully connected layer to a size that conforms to VisDA-2017 \cite{visda2017}. In other tasks, we utilize a three-layer fully connected layer structure to replace the one-layer fully connected layer structure of the original classifier. For algorithms that require a discriminator, we employ a discriminator with a three-layer fully connected layer structure. In order to eliminate the interference factors, except for the source only, all other algorithms use the same classifier and the same discriminator. We uniformly use SGD as the optimizer for training, and use $ 1\times10^{-3}$ for the learning rate of all methods. We use 64 as the batch size for training.
 
The results on VisDA-2017 are shown in Table \ref{tab:my_label}.
  We can see that our model achieves an accuracy much higher than previous methods.
  In addition, our method performs better than the source only model in all classes, whereas MCD and DANN perform worse than the source only model in some classes such as train, motorcycle, and car. Same as the last experiment, \textbf{OURS+2M} is the top-performing variant. On average, we obtain an improvement of 6.1\% compared with MCD, and 13.3\% compared with DANN. Our model also achieves the best performance in eight out of the twelve categories in this experiment.

\begin{table*}[]
    \centering
    \footnotesize
    \begin{tabular}{p{1.3cm} || p{0.35cm} p{0.35cm} p{0.3cm} p{0.3cm} p{0.35cm} p{0.35cm} p{0.35cm} p{0.4cm} p{0.35cm} p{0.35cm} p{0.35cm} p{0.45cm} || p{0.35cm}}
    \hline
       Method & { \tiny plane} & { \tiny bcycl} & { \tiny bus} &{ \tiny car} &{ \tiny horse} &{ \tiny knife} &{ \tiny mcycl} &{ \tiny person} &{ \tiny plant} &{ \tiny sktbrd} &{ \tiny train} &{ \tiny truck} &{ \tiny mean}  \\
    \hline
       {\tiny Source Only} & 30.2 & 4.3 & 27.9 & 60.6 & 31.0 & 2.1 & 82.7 & 7.4 & 67.7 & 12.9 & 79.4 & 2.3 & 40.3 \\
       {\tiny DANN \cite{ganin2015unsupervised}} & 72.3 & 53.1 & 64.7 & 31.8 & 58.2 & 14.3 & 80.7 & \textbf{60.0} & 70.0 & 41.4 & \textbf{89.7} & 20.7 & 55.9\\
       {\tiny MCD \cite{saito2018maximum}} & 82.2 & 18.7 & 86.6 & \textbf{62.1} & 73.6 & \textbf{41.0} & 89.1 & 58.9 & 80.7 & 64.3 & 74.2 & 12.5 & 63.1\\
       \hline
       {\tiny OURS} & 83.6 & 60.0 & 64.1 & 56.4 & 65.8 & 12.5 & 91.4 & 39.3 & 66.7 & 55.0 & 78.3 & 31.2 & 60.1\\
       {\tiny OURS+1M} & 85.2 & 58.5 & 76.5 & 47.1 & 73.5 & 24.7 & 89.3 & 58.9 & 75.0 & 62.2 & 80.1 & \textbf{32.0} & 63.4\\
       {\tiny OURS+2M} & \textbf{86.0} & \textbf{61.5} & \textbf{88.3} & 61.6 & \textbf{83.8} & 6.7 & \textbf{92.9} & 56.8 & \textbf{89.9} & \textbf{68.8} & 87.3 & 23.0 & \textbf{69.2}\\
    \hline
    \end{tabular}
    \caption{Results of unsupervised domain adaptation on VisDA2017 \cite{visda2017} image classification track. The accuracy is obtained by fine-tuning
    ResNet-18  \cite{he2016deep} model pre-trained on ImageNet  \cite{imagenet_cvpr09}. This task evaluates the adaptation capability from synthetic CAD model images to real-world
    MS COCO  \cite{lin2014microsoft} images. Our model achieves the best performance in most categories.} 
    \label{tab:my_label}
\end{table*}

\subsection{Ablation Study}
 We conducted ablation studies using the digital and traffic sign datasets with the same unsupervised domain adaptation setting as \textbf{ Section  \ref{sec:exp:1}}. 
 The proposed algorithm has two modules, and each module has two partial components, namely \textbf{dif-module}, \textbf{ddf-module}, \textbf{dif-MCD} and \textbf{ddf-MCD}. 
 The \textbf{dif-module} refers to the component used to learn domain invariant features in the model. 
 This is a necessary module in the algorithm and also in our baseline. 
 The \textbf{ddf-module} refers to the component used to learn domain discriminative features in the model. 
 The function of this module is to expand the discrepancy in feature distribution during training.
 The \textbf{dif-MCD} refers to the use of MCD to align the class distribution of domain invariant feature module classifiers, and the \textbf{ddf-MCD} refers to the use of MCD to align the class distribution of domain discriminative feature module classifiers.
 Therefore, we design the ablation study to test the influence of each component on the overall algorithm performance.

\begin{table}[]
    \begin{center}
    \footnotesize 
    \begin{tabular}{|p{0.77cm} p{0.77cm} p{0.5cm} p{0.5cm} | p{1.2cm} p{1.2cm} p{1.2cm} p{1.2cm} |p{0.5cm}|}
    \hline
    dif-Module & ddf-Module & dif-MCD & ddf-MCD & S$\xrightarrow{}$M & M$\xrightarrow{}$U & U$\xrightarrow{}$M & SIG$\xrightarrow{}$GTS  & Avg  \\
    \hline
    $\surd$ &  &  &  & {71.1} & {85.1} & {73.0$\,\pm\,$0.2} & {88.7} & 79.5  \\
    $\surd$ &  & $\surd$ &  & {91.4$\,\pm\,$0.2} &{97.3$\,\pm\,$0.3} & {96.8$\,\pm\,$0.1} &{90.7$\,\pm\,$0.2}  &  94.1 \\
    $\surd$ & $\surd$ &  &  & {98.9$\,\pm\,$0.1}&{95.1$\,\pm\,$0.4}&{96.1$\,\pm\,$0.2} &{91.1$\,\pm\,$0.2} & 95.3  \\
    $\surd$ & $\surd$ & $\surd$ &  & {98.3$\,\pm\,$0.1} & {97.1$\,\pm\,$0.2}  & {97.0$\,\pm\,$0.1} & {90.8$\,\pm\,$0.2}  & 95.8  \\
    $\surd$ & $\surd$ & $\surd$ & $\surd$ & \textbf{99.3$\,\pm\,$0.1} & \textbf{98.0$\,\pm\,$0.4} & \textbf{97.7$\,\pm\,$0.1} & \textbf{97.0$\,\pm\,$0.2} & \textbf{98.0} \\
    \hline
    \end{tabular}
    \caption{Ablation study of our method for unsupervised domain adaptation on  digital and traffic sign datasets.}
    \label{tab:ablation}
    \end{center}
\end{table}

 As shown in Table \ref{tab:ablation}, the performance of dual-module adversarial training has a significant improvement over using a single domain adaptation method.
 There are two most intuitive examples.
(i) The performance of $DANN$ with dual-module adversarial training is $18.5\%$ higher than the one without dual-module adversarial training. 
(ii) The performance of $DANN+MCD$ with  dual-module adversarial training is $3.9\%$ higher than the original single module.

Although the dual-module architecture costs more time in training, only one module is utilized for inference. More specifically, only G1, T1, and C1 are employed in our dual-module structure module and its variants. The introduced transformation layer T1 is a lightweight module and introduces almost negligible overhead. Thus, no matter the architecture is one module or dual-module, the inference time is almost the same. The second module in the dual-module system assists the first module to better learn the domain invariant features and lift the upper limit of the quality of the learned features.
 
\subsection{Visualization}
\begin{figure*}[t]
	\begin{minipage}[b]{0.325\columnwidth}
		\begin{center}
			\centerline{\includegraphics[width=1.0\columnwidth]{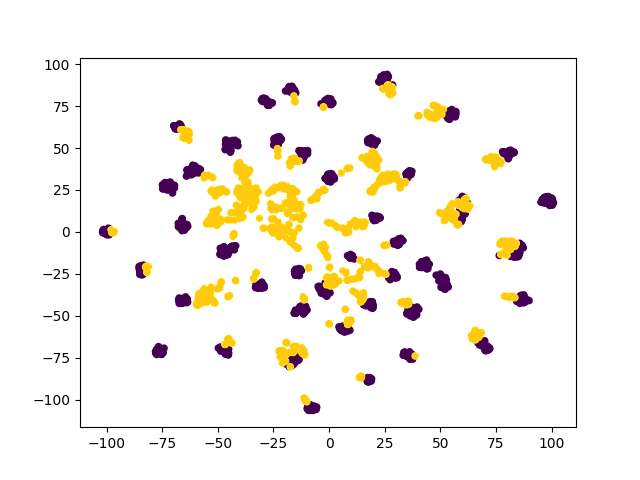}}
			{\footnotesize{Source Only}}
		\end{center}
	\end{minipage}
	\begin{minipage}[b]{0.325\columnwidth}
		\begin{center}
			\centerline{\includegraphics[width=1.0\columnwidth]{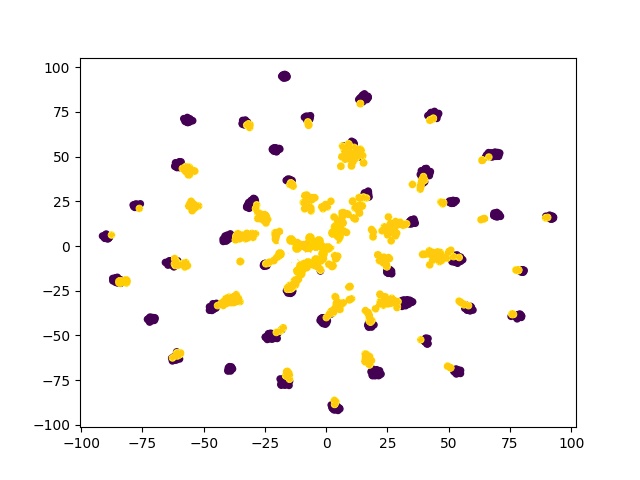}}
			{\footnotesize{Adapted: Ours}}
		\end{center}
	\end{minipage}
	\begin{minipage}[b]{0.325\columnwidth}
		\begin{center}
			\centerline{\includegraphics[width=1.0\columnwidth]{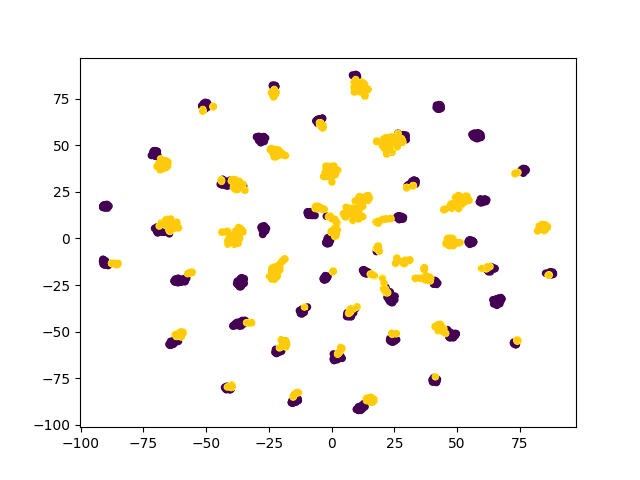}}
		    {\footnotesize{Adapted: Ours+2M}}
		\end{center}
	\end{minipage}
	\caption{\footnotesize{Visualization by using T-SNE \cite{vanDerMaaten2008}. 
        We take 2000 images from the task SYN SIGNS $\xrightarrow{}$ GTSRB. 
        Through visualization, we can easily find that our proposed method can fuse the source domain with the target domain well.
	}}
	\label{fig:fig1}
\end{figure*}

 In Figure \ref{fig:fig1}, we use T-SNE  \cite{vanDerMaaten2008} to visualize the models we trained. 
 We choose the task that transfer SYN SIGNS  \cite{978-3-319-02895-8_52} to GTSRB  \cite{Stallkamp-IJCNN-2011}. 
 SYN SIGNS is the source domain, and GTSRB is the target domain. 
 After training, we select 2000 data for visualization, where 1000 images from the source domain and 1000 images from the target domain.
 It can be clearly seen from the result that, compared to the Source Only method, our proposed model has a significant effect on reducing the domain shift, especially for the \textbf{OURS+M2} variant.
 
\section{Conclusion and Future Work}
In this paper, we have proposed a dual-module network architecture that can strongly encourage domain invariant feature learning. 
The network architecture is composed of a discriminative feature learning module and a domain invariant feature learning module. 
 We have also proposed an adversarial loss function using the difference between the feature distributions of the two modules and the similarity of their predicted results. 
 The two modules will compete with each other to maximize the difference in feature distribution.  
The proposed model employs the maximum classifier discrepancy to solve the imbalance problem of domain discriminative feature extraction in the target domain for the two modules. Extensive experiments demonstrate that the proposed method achieves state-of-the-art performance on the standard unsupervised domain adaptation benchmarks and significantly improves its performance. 

The proposed dual-module strategy can be extended to other UDA models as long as the UDA models adopt adversarial learning. 
For other UDA models, we can design dual modules like this with one module employing adversarial training for invariant features and another module exploiting regular training for discriminative features. Finally, we can adopt adversarial training between the two modules to further improve the learning ability for invariant features. We are currently working on the applications for other UDA tasks like object detection and segmentation.

\section*{Acknowledgement}

This work was partly supported in part by the Natural Sciences and Engineering Research Council of Canada (NSERC) under grant no. RGPIN-2021-04244, and the National Aeronautics and Space Administration (NASA) under grant no. 80NSSC20M0160.


\bibliographystyle{IEEEtranS}
\bibliography{mybibfile}

\end{document}